\documentclass{article}
\usepackage{amsmath,epsfig}
\usepackage[preprint]{spconfa4}

\usepackage[linesnumbered,ruled,vlined]{algorithm2e}
\SetKwInput{KwInput}{Input}                
\SetKwInput{KwOutput}{Output}              
\usepackage{multirow}
\usepackage[colorlinks,linkcolor=red,anchorcolor=blue,citecolor=blue,pagebackref=true]{hyperref}

\usepackage{float}
\usepackage{booktabs}
\usepackage{multirow} 
\usepackage{amssymb}
\usepackage{graphicx}
\usepackage{subfigure} 
\usepackage{caption} 
\usepackage{mathrsfs}

\copyrightnotice{978-1-6654-3864-3/21/\$31.00 ©2021 IEEE}

\let\OLDthebibliography\thebibliography
\renewcommand\thebibliography[1]{
  \OLDthebibliography{#1}
  \setlength{\parskip}{0pt}
  \setlength{\itemsep}{0pt plus 0.3ex}
}

\begin{document}\sloppy
\topmargin = 0mm

\def\x{{\mathbf x}}
\def\L{{\cal L}}

\title{SSDL: Self-Supervised Dictionary Learning}
%

\name{
Shuai Shao$^{a,{\ast}}$, 
Lei Xing$^{b,}$\sthanks{Shuai Shao and Lei Xing are co-first authors of this paper.}, 
Wei Yu$^c$, Rui Xu$^a$, 
Yan-Jiang Wang$^{a,{\dagger}}$, 
Bao-Di Liu$^{a,}$\sthanks{Corresponding authors: Yan-Jiang Wang and Bao-Di Liu. } 
}
\address{
$^a$College of Control Science and Engineering, China University of Petroleum (East China), 266580, China \\
$^b$College of Oceanography and Space Informatics, China University of Petroleum (East China), 266580, China \\
$^c$School of Computer Science and Technology, Harbin Institute of Technology, 264200, China\\
shuaishao@s.upc.edu.cn, \{upc\_xl, yw19960216\}@163.com, ruixu@s.upc.edu.cn, \\
yjwang@upc.edu.cn, 	thu.liubaodi@gmail.com
}
\maketitle
\begin{abstract}
The label-embedded dictionary learning (DL) algorithms generate influential dictionaries by introducing discriminative information.
However, there exists a limitation: All the label-embedded DL methods rely on the labels due that this way merely achieves ideal performances in supervised learning. While in semi-supervised and unsupervised learning, it is no longer sufficient to be effective.
Inspired by the concept of self-supervised learning 
(e.g., setting the pretext task to generate a universal model for the downstream task),
we propose a Self-Supervised Dictionary Learning (SSDL) framework to address this challenge.
Specifically, we first design a $p$-Laplacian Attention Hypergraph Learning (pAHL) block as the pretext task to generate pseudo soft labels for DL. 
Then, we adopt the pseudo labels to train a dictionary from a primary label-embedded DL method.
We evaluate our SSDL on two human activity recognition datasets.
The comparison results with other state-of-the-art methods have demonstrated the efficiency of SSDL.
\end{abstract}
\begin{keywords}
Dictionary learning, self-supervised learning, $p$-Laplacian Attention Hypergraph Learning, human activity recognition
\end{keywords}

\section{Introduction}
\label{sec:intro}
In dictionary learning, the ultimate goal is to obtain an overcomplete dictionary to represent original samples. Similar to subspace learning, the to-be-learned dictionary can be further utilized to solve different categories of problems, such as image denoising \cite{peng2014decomposable}, visual classification \cite{wang2020class}.
Many classical methods, including D-KSVD \cite{zhang2010discriminative}, LC-KSVD \cite{jiang2013label}, LEDL~\cite{shao2020label} \emph{et al.}, introduce discriminative information by adding the one-hot label matrix to the objective function. These label-embedded approaches are powerful in supervised learning, while in semi-supervised and unsupervised learning, the deficiency of labels leads to a big reduction in the effect.

Fortunately, the development of Self-Supervised Learning (SSL) provides us a novel perspective to solve this challenge. 
The core idea of SSL is to set a pretext task to generate a universal model for the downstream task.
SSL has been demonstrated to effectively address the problem caused by inadequate labeled data in the training process. 
Combined with SSL, we propose a Self-Supervised Dictionary Learning (SSDL) framework. 
Like most SSL-based methods, the critical point of the challenge is setting up an appropriate pretext task.

\begin{figure}[t]
	\begin{center}
		\includegraphics[width=0.9\linewidth]{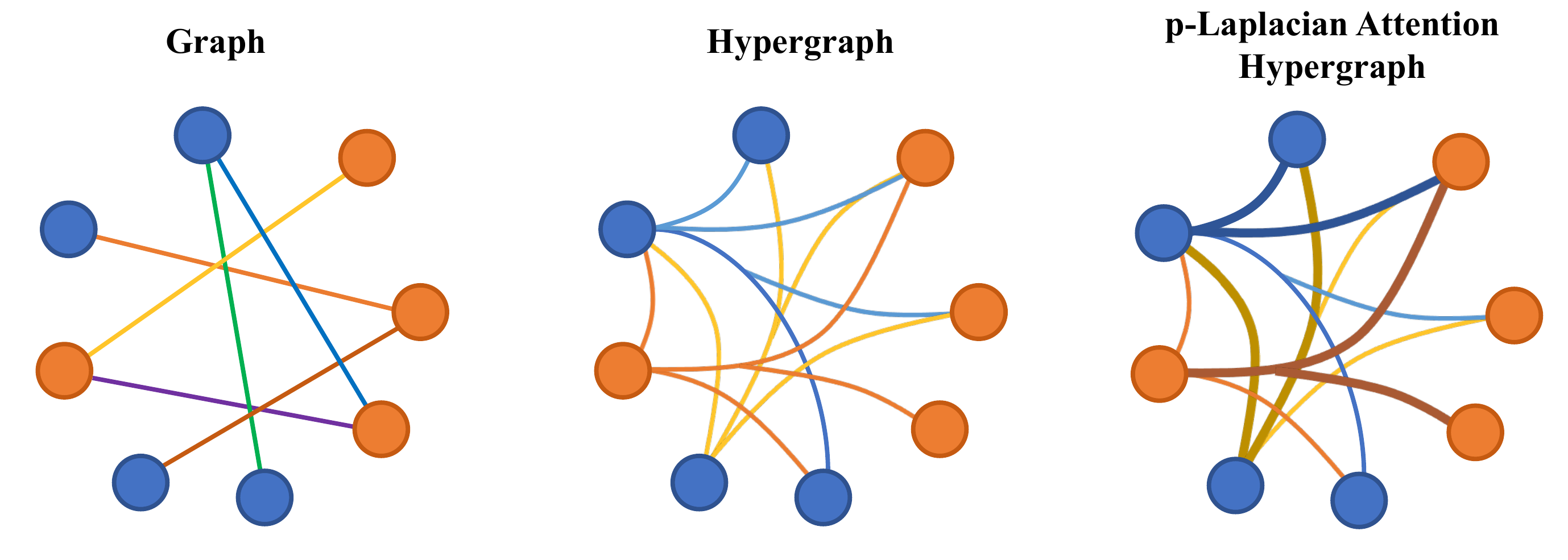}
	\end{center}
	\caption{The comparison among Graph, Hypergraph, and the proposed $p$-Laplacian Attention Hypergraph (pLA-Hypergraph). In graph structure, each edge contains two vertices. In hypergraph structure, the hyperedge is able to connect multi vertices. And in our pLA-Hypergraph, different hyperedges have different weights, which are represented by different thicknesses.}
	\label{figure: Difference_G_HG_pLHG}
\end{figure}

\begin{figure*}[t]
	\begin{center}
		\includegraphics[width=0.9\linewidth]{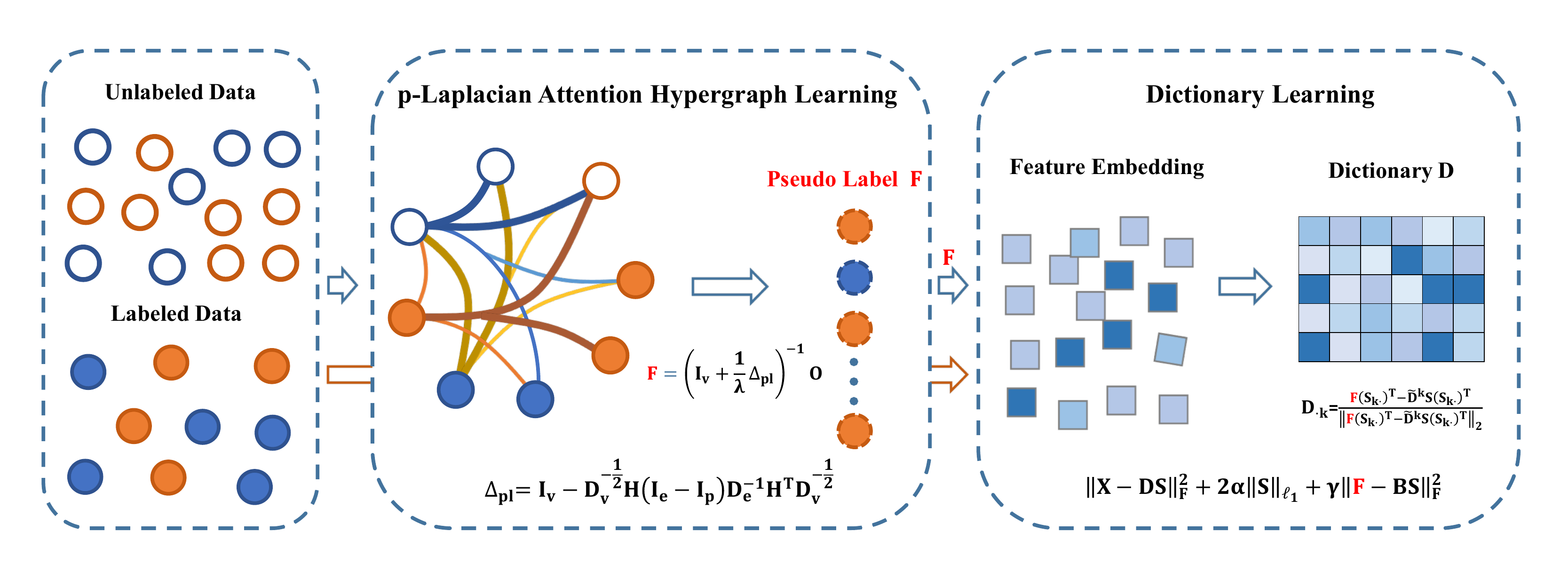}
	\end{center}
	\caption{The Self-Supervised Dictionary Learning framework. There are two steps: $i)$ Employ the pAHL block to generate pseudo label $\mathbf{F}$ for the unlabeled data. $ii)$ Embed the pseudo label into the dictionary learning model to obtain the dictionary $\mathbf{D}$. More details please refer to section~\ref{sec:Methodology}.}
	\label{figure: flowchart}
\end{figure*}

This paper proposes a $p$-Laplacian Attention Hypergraph Learning (pAHL) based pretext task to generate a pseudo label matrix and then employ it in the downstream task (e.g., DL methods).
Hypergraph learning was first proposed by Zhou \emph{et al.}\cite{zhou2007learning} in 2007. It is capable of predicting labels according to mining and aggregating high-order relations within data.
A hypergraph is composed of a vertex set and hyperedge set. Each hyperedge can connect any number of vertices. 
Compared with the simple graph, which is only able to reflect the pair-wise relations among vertices, hypergraph is more flexible and can mine deeper relations of data.

But there exists an inadequate part in traditional Laplacian-based hypergraph learning: Each hyperedge plays the equal important role in police decisions, which may lead to lose the key information sometimes. (As an example, assume that a person's weight is relevant to their diet habits and genes, but obviously, the diet habit contributes more. If we consider that these two attributes are equally important in predicting people's weight, the results would be affected.) Thus, we follow \cite{ma2018hypergraph} and introduce $p$-Laplacian regularizer to generate attention weight for each hyperedge. Note that, when $p=2$, the $p$-Laplacian regularizer is equal to the Laplacian one. We show the differences among Graph, Hypergraph, and p-Laplacian Attention Hypergraph (pLA-Hypergraph) in Figure~\ref{figure: Difference_G_HG_pLHG}.
After $p$-Laplacian Attention Hypergraph Learning, we embed the generated pseudo label matrix into a basic dictionary learning model. Figure~\ref{figure: flowchart} shows the flowchart.


In summary, the main contributions focus on:
\begin{itemize}
\item
We propose a Self-Supervised Dictionary Learning (SSDL) approach. To our best knowledge, it is the first attempt to enhance dictionary learning from the perspective of self-supervising. Specifically, we introduce $p$-Laplacian Attention Hypergraph Learning (pAHL) as the pretext task to generate a pseudo label matrix for label-embedded dictionary learning.

\item 
The proposed pAHL block is a model-agnostic method that can be employed in arbitrary standard dictionary learning to construct SSDL framework. In this paper, we just try to embed the pAHL block into a basic dictionary learning approach.

\item 
We utilize the learned dictionary in two human activity recognition tasks. The experimental results demonstrate that our SSDL is powerful, and the proposed pAHL block significantly improve the dictionary structure's performances.
\end{itemize}

\section{Methodology}
\label{sec:Methodology}

In this section, we introduce the details of the self-supervised dictionary learning algorithm. First, we introduce $p$-Laplacian based Attention Hypergraph to generate pseudo labels for the unlabeled training data. Then, we embed the pseudo label information into the standard dictionary learning framework. Figure~\ref{figure: flowchart} shows the flowchart, and Algorithm~\ref{Algorithm: SSDL} elaborates the algorithm procedure.

\subsection{Pseudo Label Generation via $p$-Laplacian Attention Hypergraph}
\label{subsec:p-Laplacian Hypergraph Learning}
\textbf{Hypergraph Construction}
A suitable hypergraph structure is beneficial to mine high-order relations among samples. Different from simple graph structure, a hypergraph $\mathcal{G}=(\mathcal{V},\mathcal{E},\mathbf{W})$ is composed of vertex set $\mathcal{V}$, hyperedge set $\mathcal{E}$, and weight matrix of hyperedge $\mathbf{W}$. The $\mathbf{W}$ is a diagonal matrix, each element denotes the weight of the corresponding hyperedge. 
Besides, there exist two degree matrices in hypergraph learning, including vertex degree matrix $\mathbf{D}_v$ and hyperedge degree matrix $\mathbf{D}_e$.
We use the incidence matrix $\mathbf{H}\in{\mathbb{R}}^{|\mathcal{V}|\times{|\mathcal{E}|}}$ to represent connections between hyperedges and vertices, and define the elements in the incidence matrix as follows:
\begin{equation}
\scriptsize
\begin{split}
        \mathbf{H}=
        \left\{\begin{array}{cc}
            {\exp \left(-dis\left(v, v_{c}\right)^{2}\right)} & {\text { if } v \in e} \\
            {0} & {\text{ o.w. }}
        \end{array}\right.
\end{split}
\label{equation: elements_in_H}
\end{equation}
where $e$ denotes one hyperedge in $\mathcal{E}$, $v$ denotes a vertex in $\mathcal{V}$. $dis$ indicates the operator to compute the distance. Following, we formulate the degree matrices as: 
\begin{equation}
\scriptsize
\begin{split}
        \delta({e})= \sum_{{v} \in \mathcal{V}} \mathbf{H}({v},{e})
\end{split}
\label{equation: vertex_degree}
\end{equation}
\begin{equation}
\scriptsize
\begin{split}
        d({v})= \sum_{{e} \in \mathcal{E}} \mathbf{W}({e} ) \mathbf{H}({v},{e})
\end{split}
\label{equation: hyperedge_degree}
\end{equation}\\
\textbf{$p$-Laplacian Attention Hypergraph Learning}
Follow \cite{gao2012laplacian}, we formulate the normalized hypergraph Laplacian regularizer as:
\begin{equation}
\scriptsize
    \begin{split}
        \mathbf{\Delta}_l=\mathbf{I}_v - \mathbf{D}_v^{-\frac{1}{2}} \mathbf{H} \mathbf{W} \mathbf{D}_e^{-1} \mathbf{H}^T \mathbf{D}_v^{-\frac{1}{2}}
    \end{split}
    \label{equation: Laplacian}
\end{equation}
where $\mathbf{I}_v \in \mathcal{R}^{|\mathcal{V}| \times |\mathcal{V}|}$ denotes the identity matrix. In most hypergraph learning tasks, the elements in $\mathbf{W}$ are set to $1$, represent that different hyperedges contribute equally for nodes aggregation. 
While in our paper, we introduce $p$-Laplacian to approximate the relations of hyperedges, to further aggregate high-order information, which can be formulated as: 
\begin{equation}
\scriptsize
    \begin{split}
        \mathbf{\Delta}_{pl}=\mathbf{I}_v - \mathbf{D}_v^{-\frac{1}{2}} \mathbf{H} \left( \mathbf{I}_e - \mathbf{L}_p \right) \mathbf{D}_e^{-1} \mathbf{H}^T \mathbf{D}_v^{-\frac{1}{2}}
    \end{split}
    \label{equation: P-Laplacian}
\end{equation}
where $\mathbf{I}_e \in \mathbb{R}^{|\mathcal{E}| \times |\mathcal{E}|}$ denotes the identity matrix.
$\mathbf{L}_p = \mathcal{Q} \Lambda \mathcal{Q}^T$.
$\mathcal{Q}= \left(q^1,q^2,\cdots,q^M \right)$ denotes the full eigenvector, and $\Lambda= \left(\Lambda^1,\Lambda^2,\cdots,\Lambda^M \right)$ denotes the corresponding eigenvalue. According to \cite{luo2010eigenvectors}, we solve the $p$-Laplacian embedding as:
\begin{equation}
\scriptsize
    \begin{split}
         &\mathop {\arg \min}\limits_{\mathcal{Q}} f_1(\mathcal{Q}) 
         =  \sum_{{m} \in \mathcal{M}} 
         \frac{\sum_{i,j \in |\mathcal{V}|} w_{ij} |q_i^m - q_j^m|^p}
         {\left\|q^m\right\|_p^p}\\
		&{\rm{s}}.t.{\kern 4pt} 
		\mathcal{Q}^T\mathcal{Q} = \mathbf{I}        
    \end{split}
    \label{equation: p-Laplacian embedding}
\end{equation}
where $w_{ij}$ is the element in $\mathbf{W}$. Here, we use the gradient method to solve Equation~\ref{equation: p-Laplacian embedding} as:
\begin{equation}
\scriptsize
    \begin{split}
        \frac{\partial f_1}{\partial q_i^m} 
        = \frac{1}{\left\|q^m\right\|_p^p}
        \left [
        \sum_j w_{ij} \phi_p \left( q_i^m - q_j^m \right) 
        - \frac{\phi_p \left( q_i^m \right)}{\left\|q^m\right\|_p^p}
        \right ]
    \end{split}
    \label{equation: p-Laplacian gradient1}
\end{equation}
where $\phi_p$ is defined that $\phi_p\left( x \right) = |x|^{p-1}sig\left( x \right)$. 
$sig$ denotes the operator to compute the negative and positive signs.
To enforce the orthogonality, we follow \cite{liu2018p} to update $\mathcal{Q}$ until convergence as:
\begin{equation}
\scriptsize
    \begin{split}
        \mathcal{Q} 
        = \mathcal{Q} 
        - \beta \left(
        \frac{\partial f_1}{\partial \mathcal{Q}}
        - \mathcal{Q} \left(
        \frac{\partial f_1}{\partial \mathcal{Q}} 
        \right)^T
        \mathcal{Q}
        \right)
    \end{split}
    \label{equation: p-Laplacian Q}
\end{equation}
where $\beta$ is the step length.
At last, we obtain the corresponding eigenvalue as :
\begin{equation}
\scriptsize
    \begin{split}
        \Lambda^m
         =  \frac{\sum_{i,j \in |\mathcal{V}|} w_{ij} |q_i^m - q_j^m|^p}
         {\left\|q^m\right\|_p^p}
    \end{split}
    \label{equation: p-Laplacian lambda}
\end{equation}\\
\textbf{Pseudo Label Generation}
Assume parts of training data have labels, define initial label embedding matrix as $\mathbf{O} \in \mathbb{R}^{C \times N}$, where $C$ denotes the total number of classes. For labeled samples, $\mathbf{O}_{ij}$ is $1$ if the $j$-th sample belongs to the $i$-th class, and it is $0$ otherwise. For unlabeled samples, we set all elements to $0.5$. 
We formulate the objective function as:
\begin{equation}
\scriptsize
\begin{split}
         &\mathop {\arg \min}\limits_{\mathbf{F}} f_2(\mathbf{F}) 
         = \text{tr} \left( \Delta_{pl} \mathbf{F}^T \mathbf{F} \right)
         + \lambda \left\| \mathbf{F} - \mathbf{O} \right\|_F^2\\
\end{split}
\label{equation: Pseudo Label Obj}
\end{equation}
where $\lambda$ is the parameter to balance the objective function. According to \cite{zhou2007learning}, we directly obtain the pseudo label as:
\begin{equation}
\scriptsize
\begin{split}
        \mathbf{F} = \left(\mathbf{I}_v + \frac{1}{\lambda} \mathbf{\Delta}_{pl} \right) ^{-1} \mathbf{O}
\end{split}
\label{equation: Pseudo Label}
\end{equation}
where $\mathbf{F} \in \mathbb{R}^{C \times N}$ is the predicted pseudo label matrix. Unlike the one-hot truth label matrix, the $\mathbf{F}$ is soft.

\subsection{Self-Supervised Dictionary Learning}
\label{sec: Self-Supervised Dictionary Learning}
The above section shows that the learned pseudo label information only relies on the hypergraph structure. That is to say, the proposed $p$-Laplacian Attention Hypergraph Learning (pAHL) is a model-agnostic approach, which can be embedded into any dictionary learning framework. Here, we just introduce the pAHL block into a standard dictionary learning. The objective function can be formulated as:
\begin{equation}
\scriptsize
\begin{split}
        &\mathop {\arg \min}\limits_{\mathbf{D}, \mathbf{S},\mathbf{B}}     f_3(\mathbf{D},\mathbf{S},\mathbf{B})\\
		&=\left\| \mathbf{X} - \mathbf{D}\mathbf{S} \right\|_F^2 
		+ 2\alpha \left\| \mathbf{S} \right\|_{\ell_1} 
		+ \gamma {\kern 2pt} \left\| \mathbf{F} - \mathbf{B}\mathbf{S} \right\|_F^2 \\
		&{\kern 5pt}{\rm{s}}.t.\left\| {{{\bf{d}}_{ \bullet k}}} \right\|_2^2 \le 1,
		{\kern 5pt}\left\| {{{\bf{b}}_{ \bullet k}}} \right\|_2^2 \le 1 {\kern 4pt} \left( {k = 1,2, \cdots K} \right)\\
\end{split}
\label{equation: Objective_function}
\end{equation}
where $\mathbf{X}=[{\mathbf{x}}_1,{\mathbf{x}}_2,\dots,{\mathbf{x}}_{N}] \in \mathbb{R}^{dim \times N}$ denotes the training data, ${\mathbf{x}}_i$ ($i = 1, 2, \dots$) denotes the feature embedding of the $i$-${th}$ sample, $dim$ denotes the dimension size of each sample, $N$ is the number of training samples.
$\mathbf{D} \in \mathbb{R}^{dim \times K}$ represents the to-be-learned dictionary, $K$ is the dictionary base size. 
$\mathbf{B} \in \mathbb{R}^{C \times K}$ represents the to-be-learned classifier, $C$ denotes the class number.
$\mathbf{S} \in \mathbb{R}^{K \times N}$ denotes the sparse codes for dictionary.
$\alpha$ and $\gamma$ are the positive scalar constants.

We alternate update $\mathbf{S}$, $\mathbf{D}$ and $\mathbf{B}$ until the objective function doesn’t descend. $\mathbf{S}$ can be solved as:
\begin{equation}
\scriptsize
\begin{split}
        \mathbf{S}_{kn}=
            {\frac{max\left(\mathcal{J},\alpha \right) + min\left(\mathcal{J},\alpha \right)}
            {\left(\mathbf{D}^T\mathbf{D}
            +\gamma {\kern 2pt} \mathbf{B}^T\mathbf{B}\right)_{kk}}}
\end{split}
\label{equation: optimization_UpdateS_Skn_1}
\end{equation}
where 
\begin{equation}
\scriptsize
\begin{split}
   \mathcal{J}&=\left(\mathbf{D}^T\mathbf{X}+\gamma \mathbf{B}^T \mathbf{F} \right)_{kn}
    -\sum_{l=1,l\neq k}^{K} \left(\mathbf{D}^T\mathbf{D}
    +\gamma \mathbf{B}^T\mathbf{B} \right)_{kl}\mathbf{S}_{ln}
\end{split}
\label{equation: optimization_UpdateS_J}
\end{equation}
Then we introduce BCD~\cite{liu2014blockwise} to update $\mathbf{B}$ and $\mathbf{D}$ as:
\begin{equation}
\scriptsize
\begin{split}
            \mathbf{D}_{\bullet k} 
            = \frac{\mathbf{X}\left(\mathbf{S}_{k \bullet} \right)^T
            -\mathbf{\tilde{D}}^k \mathbf{S} \left(\mathbf{S}_{k \bullet} \right)^T}
            {\| \mathbf{X}\left(\mathbf{S}_{k \bullet} \right)^T
            -\mathbf{\tilde{D}}^k \mathbf{S} \left(\mathbf{S}_{k \bullet} \right)^T \|_2}   
\end{split}
\label{equation: optimization_UpdateD}
\end{equation}
\begin{equation}
\scriptsize
\begin{split}
            \mathbf{B}_{\bullet k} 
            = \frac{\mathbf{F}\left(\mathbf{S}_{k \bullet} \right)^T
            -\mathbf{\tilde{B}}^k \mathbf{S} \left(\mathbf{S}_{k \bullet} \right)^T}
            {\| \mathbf{F}\left(\mathbf{S}_{k \bullet} \right)^T
            -\mathbf{\tilde{B}}^k \mathbf{S} \left(\mathbf{S}_{k \bullet} \right)^T \|_2}     
\end{split}
\label{equation: optimization_UpdateB}
\end{equation}
where $ \mathbf{\tilde{D}}=
        \left\{ \begin{array}{cc}
             {\mathbf{D}_{\bullet p}}  & {p \neq{k}}  \\
             {\mathbf{0}} & {p = k}
         \end{array}\right.$, 
         $ \mathbf{\tilde{B}}=
        \left\{ \begin{array}{cc}
             {\mathbf{B}_{\bullet p}}  & {p \neq{k}}  \\
             {\mathbf{0}} & {p = k}
         \end{array}\right.$, $\mathbf{0}$ denotes zero matrix.
We conduct the Self-Supervised Dictionary Learning method in Algorithm~\ref{Algorithm: SSDL}. 
\begin{algorithm}[t]
\DontPrintSemicolon
  
  \KwInput{$\mathbf{X} \in \mathbb{R}^{dim \times N}$}
  \KwOutput{$\mathbf{D} \in \mathbb{R}^{dim \times K}$, $\mathbf{S} \in \mathbb{R}^{K \times N}$}
    Construct hypergraph $\mathbf{H}$ by \textbf{Equation \ref{equation: elements_in_H}}.\\

    \While{i \textless maxitem}
    {
        Solve $p$-Laplacian embedding, update eigenvector $\mathcal{Q}$ by \textbf{Equation \ref{equation: p-Laplacian gradient1}, \ref{equation: p-Laplacian Q}}.\\
        Update eigenvalue $\Lambda$ by \textbf{Equation \ref{equation: p-Laplacian lambda}}.\\
    }
    
    Obtain $p$-Laplacian-based attention hypergraph regularizer $\mathbf{\Delta}_{pl}$ by \textbf{Equation \ref{equation: P-Laplacian}}.\\
    
    Generate pseudo label $\mathbf{F}$ by \textbf{Equation \ref{equation: Pseudo Label Obj}, \ref{equation: Pseudo Label}}.\\

    \While{j \textless maxitem}
    {
        Update sparse codes $\mathbf{S}$ by \textbf{Equation \ref{equation: optimization_UpdateS_Skn_1},\ref{equation: optimization_UpdateS_J}}.\\
        Update dictionary $\mathbf{D}$ by \textbf{Equation \ref{equation: optimization_UpdateD}}.\\
        Update classifier $\mathbf{B}$ by \textbf{Equation \ref{equation: optimization_UpdateB}}.\\
    }
    
	\caption{Self-Supervised Dictionary Learning}
	\label{Algorithm: SSDL}
\end{algorithm}

\section{Experiment}
\label{sec: experiment}
Dictionary learning has been widely applied in many fields. Here we evaluate the learned dictionary in human activity recognition tasks. There are two datasets, including Stanford 40 Actions (Stanford40) \cite{yao2011human} dataset and UIUC Sports Event (UIUC-SE) \cite{li2007and} dataset. 
We first introduce the experimental setup. Then compare the proposed SSDL with state-of-the-art methods. 
Next, we try to embed the proposed pAHL block into other classical methods to evaluate the model-agnostic ability. 
Following, we conduct ablation studies to analyze our method. At last, we discuss something about the pretext task.

\subsection{Experimental Setup}
\label{sec: exp_set_up}
For all the datasets, we employ standard Resnet to extract feature embedding with $2,048$ dimensions, select $70\%$ for training, the rest for testing, and only $40\%$ training data has labels.
For the $p$ and $\lambda$ in pretext task, they play the key roles to obtain a suitable pseudo label matrix for dictionary learning. 
We fix them to $1.8$, $0.1$ for Stanford40, and $2.2$, $0.1$ for UIUC-SE. There is a trick to tune the two parameters, for more details, please refer to section~\ref{sec: Ablation Studies}.
In dictionary learning, we set the dictionary size $K$ to half the number of training samples for the two datasets, and $\alpha=2^{-14}$, $\gamma=2^{-12}$ for Stanford40 dataset, $\alpha=2^{-12}$, $\gamma=2^{-12}$ for UIUC-SE dataset. The details are also discussed in section~\ref{sec: Ablation Studies}.

\begin{table}[t]
        \caption{Recognition results with $40\%$ label rates.}
        \label{table: Recognition_results_40}
        \begin{center}
            \begin{tabular}{lcc}
                \toprule
                \textbf{Methods$\backslash$Datasets}            & \textbf{Stanford40}    & \textbf{UIUC-SE}          \\
                \midrule
                SRC (TPAMI \cite{wright2009robust}, 2009)       & 66.0$\%$               & 88.4$\%$      \\
                CRC (ICCV \cite{zhang2011sparse}, 2011)         & 70.1$\%$               & 94.2$\%$      \\
                NRC (PR \cite{xu2019sparse}, 2019)              & 67.7$\%$               & 89.7$\%$      \\
                SLRC (TPAMI \cite{deng2018face}, 2018)          & 65.3$\%$               & 93.4$\%$       \\
                Euler-SRC (AAAI \cite{liu2018euler}, 2018)      & 66.9$\%$               & 90.2$\%$      \\
                \midrule                
                ADDL (TNNLS \cite{zhang2018jointly}, 2018)      & 74.8$\%$               & 95.7$\%$       \\
                FDDL (ICCV \cite{yang2011fisher}, 2011)         & 73.3$\%$               & 94.2$\%$     \\
                LC-KSVD (TPAMI \cite{jiang2013label}, 2013)     & 67.7$\%$               & 89.1$\%$      \\
                LC-PDL (IJCAI \cite{zhang2019scalable}, 2019)   & 73.3$\%$               & 91.3$\%$      \\
                LEDL (NC \cite{shao2020label}, 2020)            & 72.9$\%$               & 91.8$\%$      \\
                CDLF (SP \cite{wang2020class}, 2020)            & 72.7$\%$               & 92.4$\%$       \\
                \midrule
                \textbf{SSDL}                                   & \textbf{75.9$\%$}      & \textbf{96.4$\%$}  \\
                \bottomrule
            \end{tabular}        
        \end{center}
\end{table}

\begin{figure}[t]
	\begin{center}
		\includegraphics[width=0.9\linewidth]{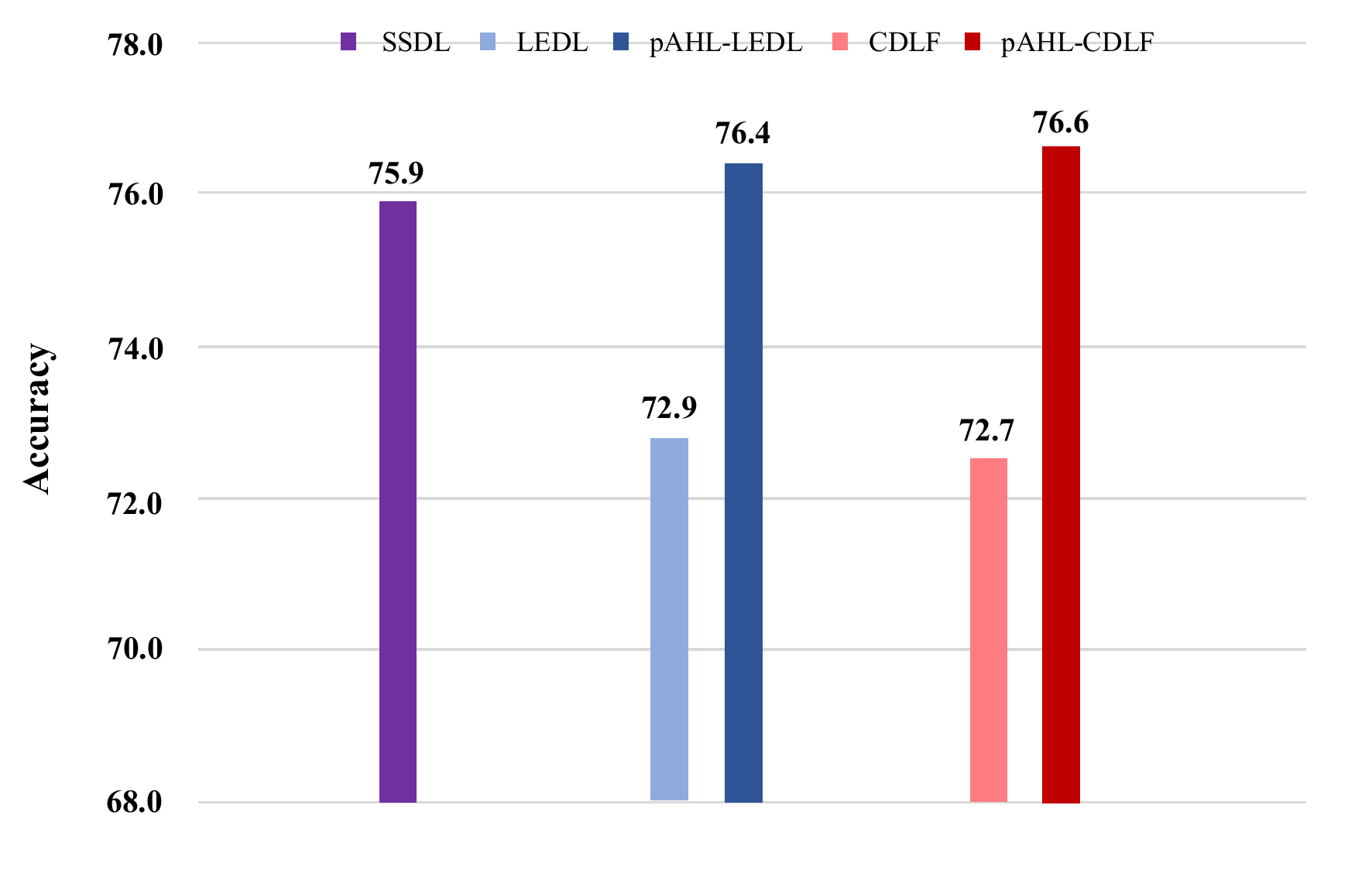}
	\end{center}
	\caption{Comparison results about pAHL-LEDL and pAHL-CDLF on Stanford40 dataset with $40\%$ label rates.}
	\label{figure: pAHL_LEDL_CDLF}
\end{figure}

\begin{figure*}[t]
    \subfigure[]{
        \begin{minipage}[t]{0.25\linewidth}
        	\begin{center}
        		\includegraphics[width=1\linewidth]{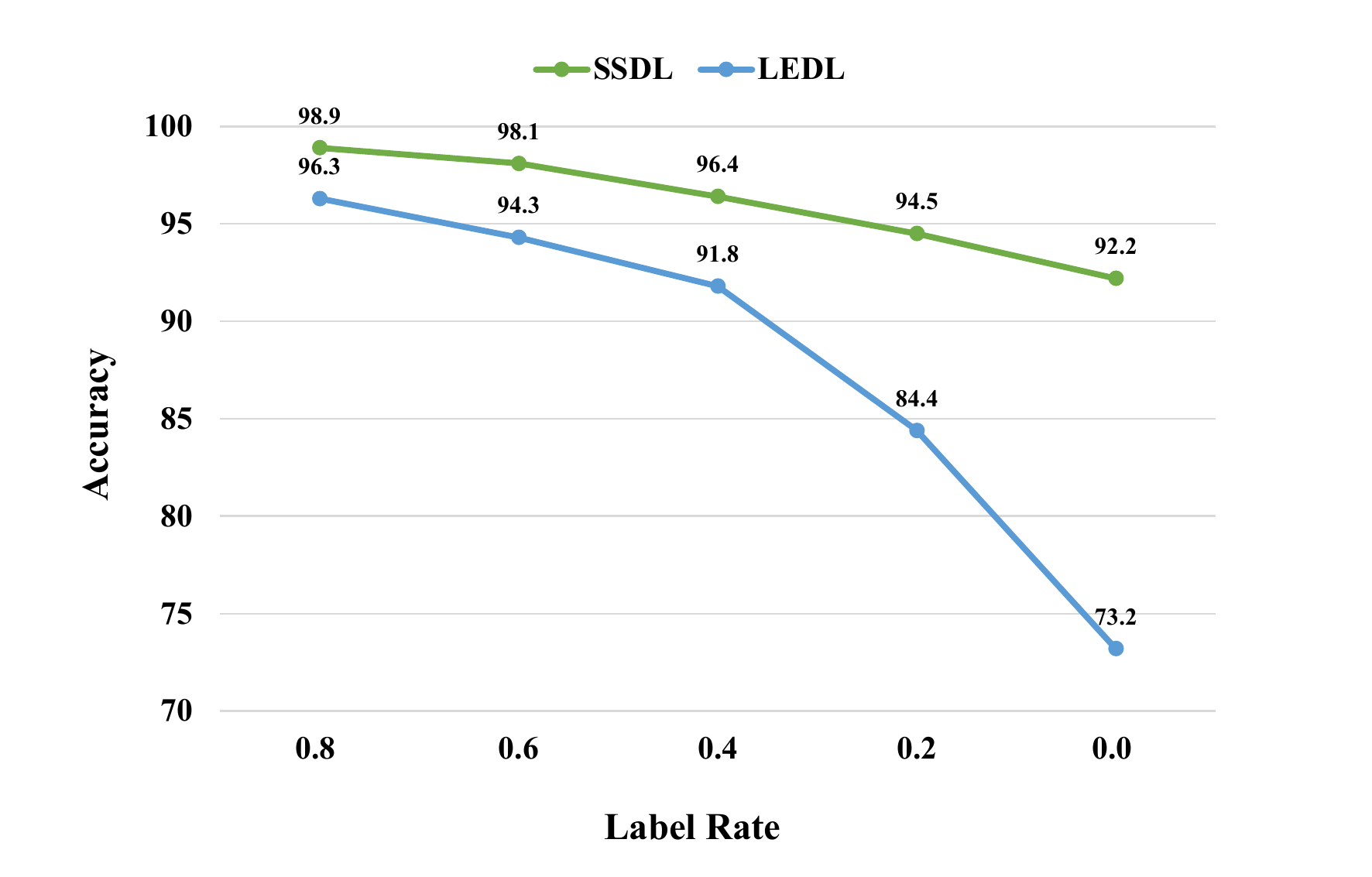}
        	\end{center}
        	\label{figure: label_rate}
        \end{minipage}%
    }%
    \subfigure[]{
        \begin{minipage}[t]{0.25\linewidth}
        	\begin{center}
        		\includegraphics[width=1\linewidth]{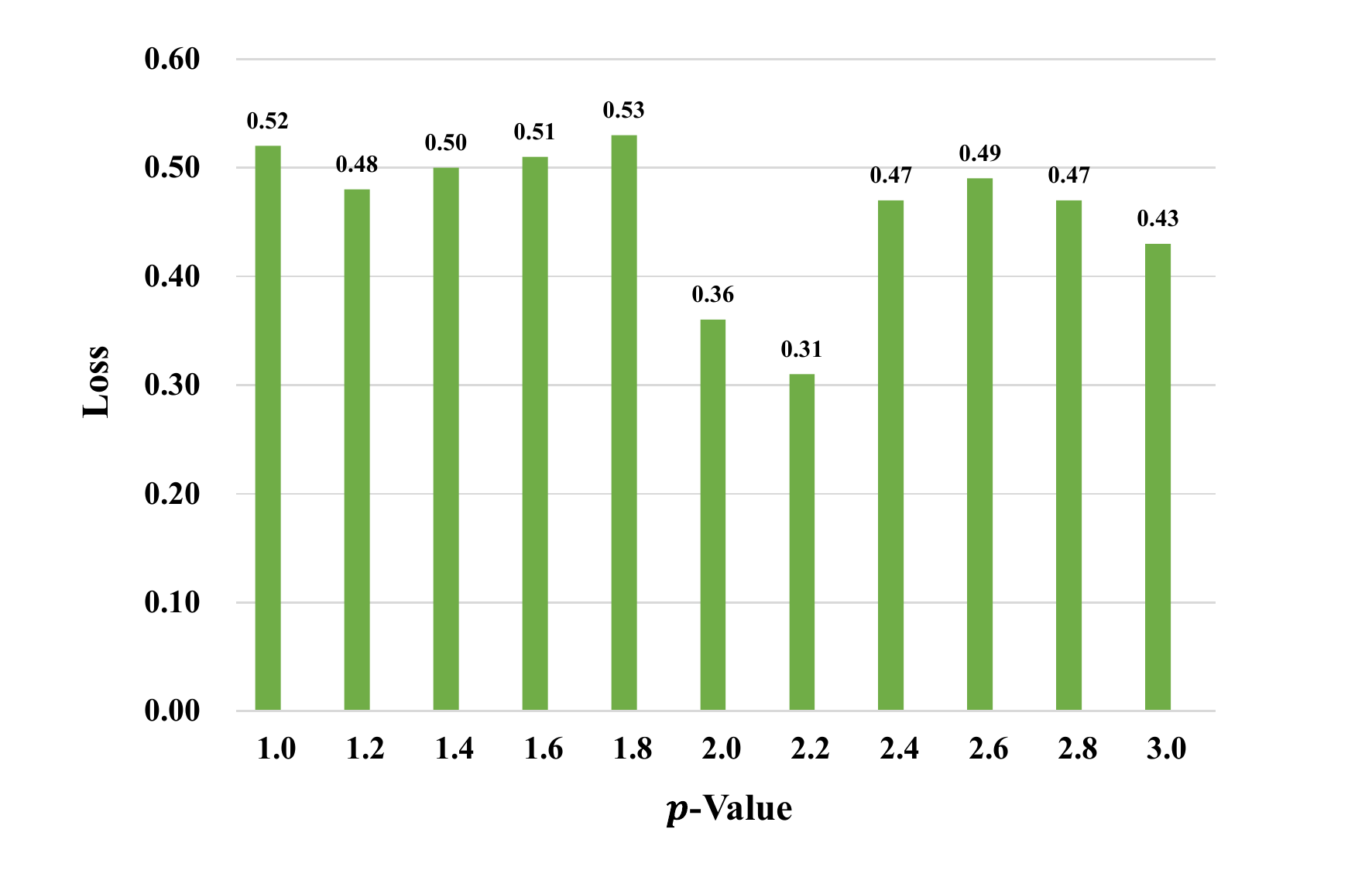}
        	\end{center}
        	\label{figure: p_parameter}
        \end{minipage}%
    }%
    \subfigure[]{
        \begin{minipage}[t]{0.25\linewidth}
        	\begin{center}
        		\includegraphics[width=1\linewidth]{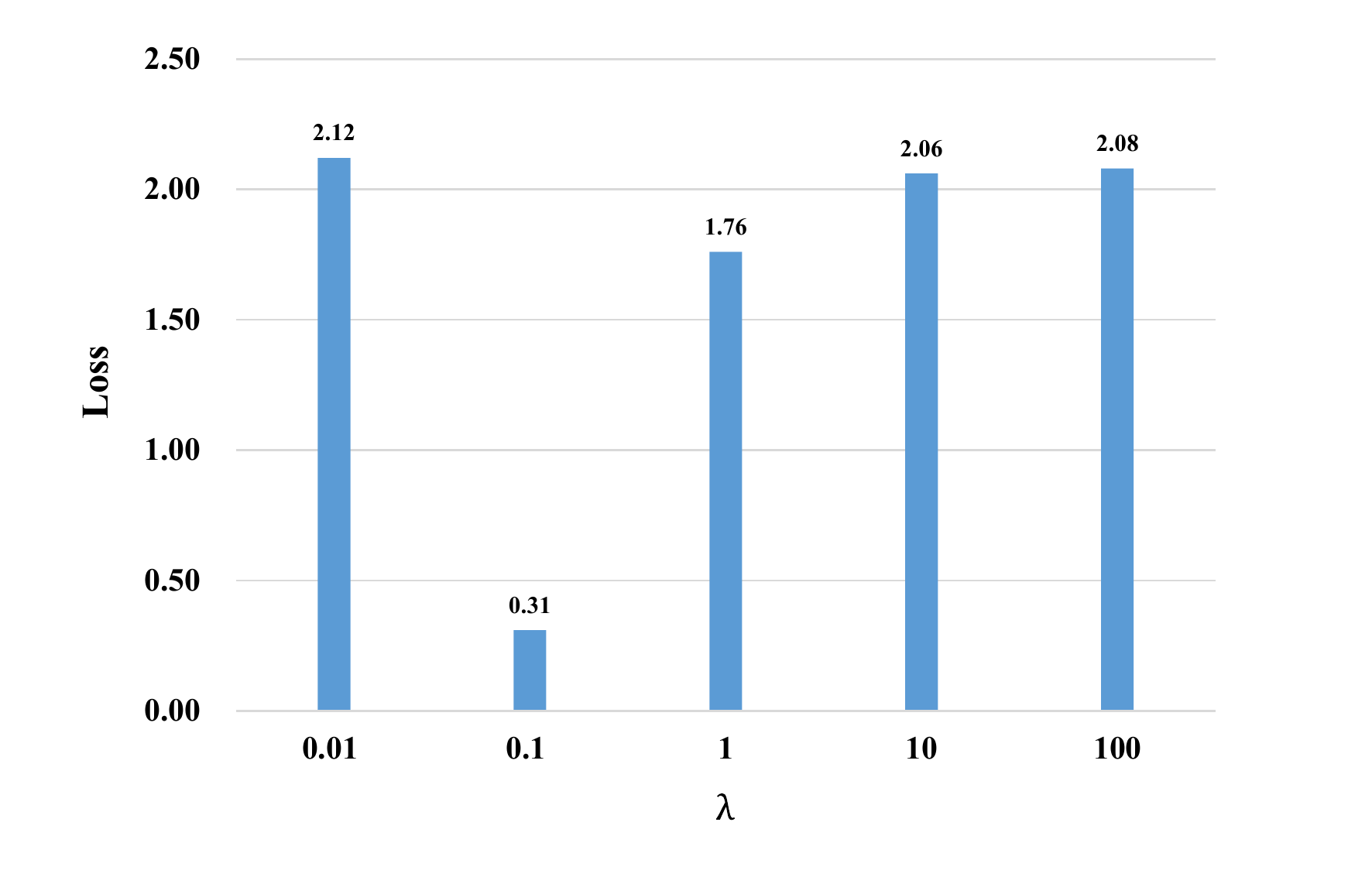}
        	\end{center}
        	\label{figure: lambda_parameter}
        \end{minipage}%
    }%
    \subfigure[]{
    \begin{minipage}[t]{0.25\linewidth}
    	\begin{center}
    		\includegraphics[width=1\linewidth]{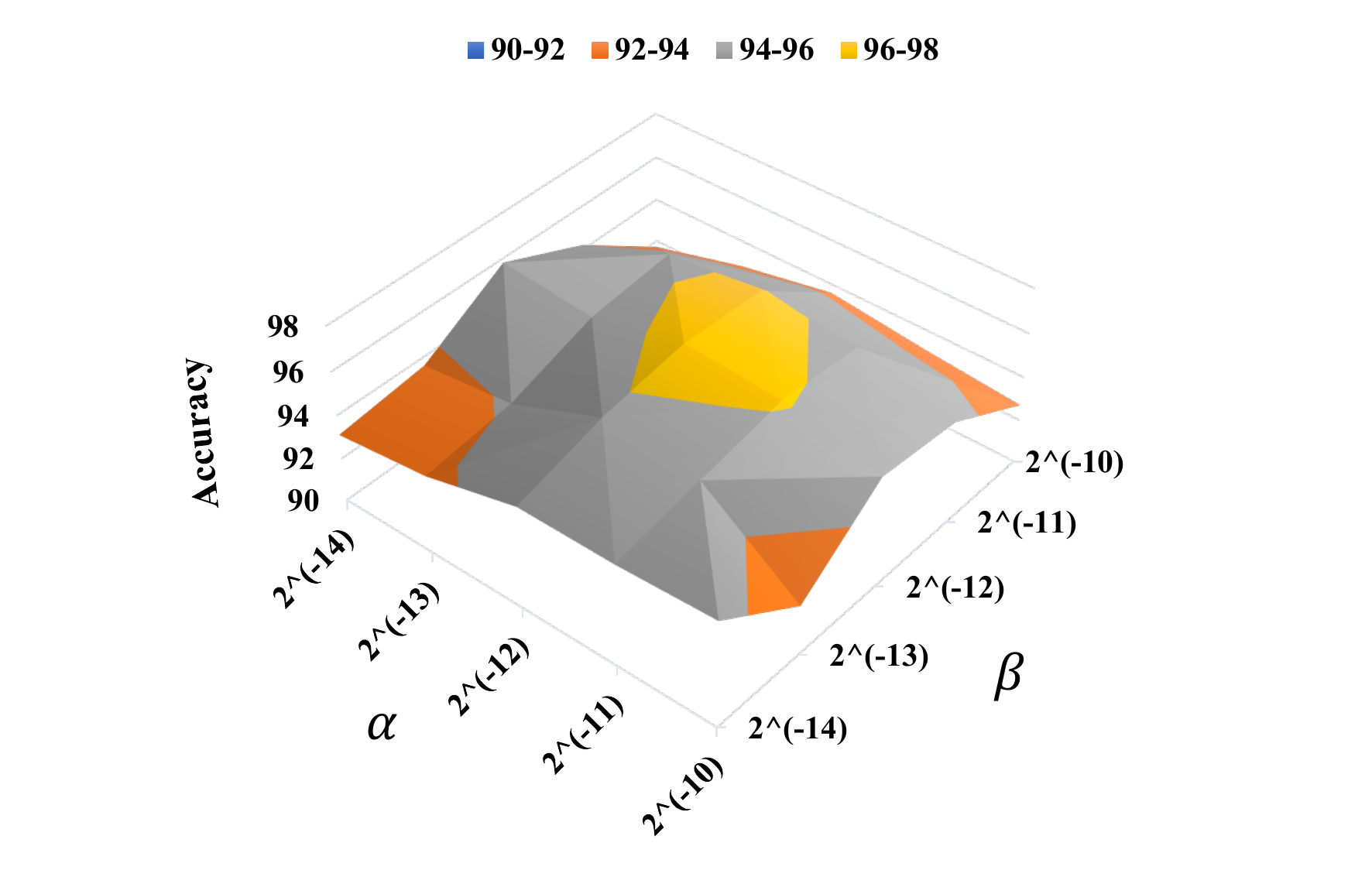}
    	\end{center}
    	\label{figure: alpha_beta_parameter}
    \end{minipage}%
    }%
    \caption{Ablation studies}
\end{figure*}

\subsection{Experimental Results}
\label{sec: exp_result}
We compare our SSDL with other state-of-the-art methods. We split these approaches into two categories, which are separated by horizontal lines in Table~\ref{table: Recognition_results_40}: 
$i)$ Traditional machine learning methods (directly use the testing samples to fit the training samples), including SRC \cite{wright2009robust}, CRC \cite{zhang2011sparse}, NRC \cite{xu2019sparse}, SLRC \cite{deng2018face} and Euler-SRC \cite{liu2018euler}. 
$ii)$ Dictionary learning methods, including ADDL \cite{zhang2018jointly}, FDDL \cite{yang2011fisher}, LC-KSVD \cite{jiang2013label},  LC-PDL \cite{zhang2019scalable}, LEDL \cite{shao2020label}, CDLF \cite{wang2020class}.
We show the recognition results with $40\%$ labeled training data in Table~\ref{table: Recognition_results_40} and have the following observations.

From Table~\ref{table: Recognition_results_40}, we can see that our SSDL can outperform all other methods at least $1.1\%$ and $0.7\%$
on the Stanford40 and UIUC-SE datasets, respectively. 
Compared with the traditional methods, our SSDL has significant improvements, but we need to consume more resources when training the dictionary. 
Compared with other state-of-the-art dictionary learning based approaches, SSDL has at least $0.7\%$ improvement. 
For the label-embedded dictionary learning methods (LC-KSVD, LC-PDL, LEDL, CDLF), SSDL's recognition accuracies can exceed them at least $2.6\%$. This phenomenon has demonstrated the efficiency of our method to some extent.

However, our SSDL just embeds the pAHL based pretext task into a basic dictionary learning model. As mentioned in section~\ref{sec:intro}, the pAHL block is a model-agnostic method that can be embedded into any standard dictionary learning algorithm, such as LC-KSVD, LC-PDL, LEDL, CDLF. That is to say, we may achieve higher recognition accuracies if we try to embed our pAHL block into these models. 
To evaluate this statement, we expand pAHL block to LEDL and CDLF on the Stanford40 dataset. The results are shown in Figure~\ref{figure: pAHL_LEDL_CDLF}.  Obviously see that, compared with original methods, the pAHL-embedded LEDL and CDLF can achieve more powerful performances than SSDL.

\subsection{Ablation Studies}
\label{sec: Ablation Studies}
The SSDL approach has achieved outstanding performance. It is interesting to recognize what are the factors affecting the experimental results. For this purpose, we design two ablation studies to discuss the proposed SSDL method.


$i)$ One of our approach's main contributions is to reduce the dependence on labeled data for dictionary learning. Thus, we design an ablation study on the UIUC-SE dataset to observe the effect of label rates. From Figure~\ref{figure: label_rate}, we can see that, with the decrease of label rates, the performances of the two methods are decreasing, but our method is much slower than the other one.

$ii)$ There are mainly four parameters ($p$, $\lambda$, $\alpha$, $\gamma$) influence the results. We set all the evaluated experiments to $40\%$ label rate on the UIUC-SE dataset.
Here, we first discuss the $p$ and $\lambda$ in the pretext task. 
We adjust $p$ and $\lambda$ to obtain a pseudo label matrix. Usually, we fine-tune the two parameters according to the final results (as an example, in our paper, we can adjust the two parameters by the recognition accuracy). Here, we give a trick to easier ensure the two optimal parameters. Specifically, we first use the training data to generate a model with $p$ and $\lambda$. Then employ the training model to compute the cross-entropy loss of testing data. At last, adjust the parameters until achieving the minimum loss. 
The influence of $p$ and $\lambda$ are separately shown in Figure \ref{figure: p_parameter}, \ref{figure: lambda_parameter}. The y-axis denotes the testing data's loss. We  obtain the minimum loss near $p=2.2$ and $\lambda=0.1$.
For $\alpha$ and $\gamma$, they interact with each other. Thus we explore the impact of these two parameters simultaneously. Figure~\ref{figure: alpha_beta_parameter} shows the experimental results. The proposed SSDL approach is not sensitive to these two parameters.

\subsection{Pretext Task}
\label{sec: Pretext Task}
In our framework, we set our proposed pAHL as the pretext task. Actually, it is flexible to select other methods, such as GL \cite{zhou2003learning}, HL \cite{zhou2007learning}, HL-W \cite{gao20123}, DHSL \cite{zhang2018dynamic}, to predict the pseudo label for dictionary learning. We employ the cross entropy loss to describe the influence. Results are shown in Table~\ref{table: loss}. Obviously see that, our pAHL is able to get better performance than GL, HL, and HL-W, but obtain similar results with DHSL.

\begin{table}[t]
        \caption{Cross-entropy loss on pretext task with $40\%$ label rates.}
        \label{table: loss}
        \begin{center}
            \begin{tabular}{lcc}
                \toprule
                \textbf{Methods$\backslash$Datasets}            & \textbf{Stanford40}    & \textbf{UIUC-SE}          \\
                \midrule
                GL (NIPS \cite{zhou2003learning}, 2003)         & 0.47              & 0.51      \\
                HL (NIPS \cite{zhou2007learning}, 2007)         & 0.64              & 0.42      \\
                HL-W (TIP \cite{gao20123}, 2012)                & 0.51              & 0.36      \\
                DHSL (IJCAI \cite{zhang2018dynamic}, 2018)      & 0.49              & \textbf{0.27}       \\
                \midrule
                \textbf{pAHL}                                   & \textbf{0.43}      & 0.31  \\
                \bottomrule
            \end{tabular}        
        \end{center}
\end{table}



\section{Conclusion}
Label-embedded dictionary learning is a typical technology in machine learning. However, limited to introducing the label information, this category of approaches is only appliable in supervised learning. Inspired by the self-supervised idea, we propose a self-supervised dictionary learning method to expand label-embedded dictionary learning to semi-supervised and unsupervised learning. To our best knowledge, this is the first attempt to solve this dictionary learning challenge from the self-supervised perspective. Experimental results have demonstrated the efficiency of our method.

\section{Acknowledgements}
The paper was supported by the National Natural Science Foundation of China (Grant No. 62072468), the Natural Science Foundation of Shandong Province, China (Grant No. ZR2019MF073, ZR2018MF017), the Open Research Fund from Shandong Provincial Key Laboratory of Computer Network (No. SDKLCN-2018-01), Qingdao Science and Technology Project (No. 17-1-1-8-jch), the Fundamental Research Funds for the Central Universities, China University of Petroleum (East China) (Grant No. 20CX05001A), the Major Scientific and Technological Projects of CNPC (No. ZD2019-183-008), and the Creative Research Team of Young Scholars at Universities in Shandong Province (No.2019KJN019).
\bibliographystyle{IEEEbib}
\bibliography{icme2021template}

\end{document}